\documentclass[10pt,twocolumn,letterpaper]{article}

\usepackage[pagenumbers]{iccv} %
\usepackage{multirow}
\usepackage{amssymb}
\usepackage{pifont}
\usepackage{balance}
\usepackage[table, dvipsnames]{xcolor}

\def\x{\mathbf{x}}
\def\p{\mathbf{p}}
\def\y{\mathbf{y}}
\def\xi{\x_i}
\def\pi{\p_i}
\def\ci{\mathbf{c}_i}

\def\pj{\p_j}
\def\cj{\mathbf{c}_j}
\def\yj{\y_j}
\def\ygsj{\y_j^{GS}}
\def\yhatj{\hat{\y}_j}
\def\xref{\x_r}

\def\pc{\mathbf{\Phi}}
\def\gsmodel{\mathcal{G}}
\def\difix{\mathcal{E}}

\definecolor{iccvblue}{rgb}{0.21,0.49,0.74}
\usepackage[pagebackref,breaklinks,colorlinks,allcolors=iccvblue]{hyperref}

\def\paperID{*****} %
\def\confName{ICCV}
\def\confYear{2025}

\title{Hybrid Gaussian Splatting for Novel Urban View Synthesis}

\author{
Mohamed Omran* \quad Farhad Zanjani* \quad Davide Abati* \quad Jens Petersen \quad Amirhossein Habibian
\vspace{0.5em}\\
Qualcomm AI Research$^1$\\
\small{\texttt{\{momran,fzanjani,dabati,jpeterse,ahabibia\}@qti.qualcomm.com}}
}

\newcommand\blfootnote[1]{%
  \begingroup
  \renewcommand\thefootnote{}\footnote{#1}%
  \addtocounter{footnote}{-1}%
  \endgroup
}

\DeclareMathOperator*{\argmin}{arg\,min}

\begin{document}
\maketitle
\blfootnote{$^1$Qualcomm AI Research is an initiative of Qualcomm Technologies, Inc.}

\begin{abstract}

This paper describes the Qualcomm AI Research solution to the RealADSim-NVS challenge, hosted at the RealADSim Workshop at ICCV 2025.
The challenge concerns novel view synthesis in street scenes, and participants are required to generate, starting from car-centric frames captured during some training traversals, renders of the same urban environment as viewed from a different traversal (e.g. different street lane or car direction).
Our solution is inspired by hybrid methods in scene generation and generative simulators merging gaussian splatting and diffusion models, and it is composed of two stages: First, we fit a 3D reconstruction of the scene and render novel views as seen from the target cameras. 
Then, we enhance the resulting frames with a dedicated single-step diffusion model.
We discuss specific choices made in the initialization of gaussian primitives as well as the finetuning of the enhancer model and its training data curation.
We report the performance of our model design and we ablate its components in terms of novel view quality as measured by PSNR, SSIM and LPIPS.
On the public leaderboard reporting test results, our proposal reaches an aggregated score of 0.432, achieving the second place overall.
\end{abstract}

\section{Introduction}
Novel View Synthesis (NVS) refers to the problem of generating, given a set of images of an object or a scene captured from several source viewpoints, other renders of the same subject or environment captured from different views.
This task plays a crucial role in several computer graphics and computer vision applications.
Among them, NVS has clear implications in autonomous driving: indeed, as testing driving agents in real-world comes with high safety concerns and prohibitive costs, relying on simulation is often preferable.
In this respect, NVS represents a core feature of modern generative simulators~\cite{hugsim}.
Think, for instance, at the necessity of evaluating the performance of a planner algorithm in a prerecorded urban scene: as the agent goes through the environment, its observations need to be rendered according to its own driving trajectory, that might differ from the one followed to collect the scene.

Our solution to the NVS problem is inspired by the recent trend of hybrid models merging 3D reconstruction and diffusion models.
Examples of these hybrid approaches can be found, besides in novel view synthesis~\cite{difix}, in related tasks such as urban scene generation~\cite{magicdrive3d,drivedreamer4d,dreamdrive,stereoforcing,stag1}.

In this report, we describe our approach along with some findings we made along the way.
Our pipeline is composed of two main steps.
First, we use the source images and viewpoints to reconstruct the 3D geometry of the scene with 3D Gaussian Splatting~\cite{3dgs}.
We then initialize the novel views by rendering from the reconstructed environment, and we postprocess the resulting frame with a dedicated enhancer model~\cite{difix}, that removes rendering artifacts.
We illustrate the importance of point cloud initialization of the 3DGS using Structure-From-Motion, and we describe our strategy to curate the training data for finetuning the enhancer model.
Overall, our proposal achieved an aggregated score of 0.432, worth the $2^{nd}$ place among 28 submissions.

\section{NVS Problem formulation}
\begin{figure*}[t]
\centering
\includegraphics[width=0.9\textwidth]{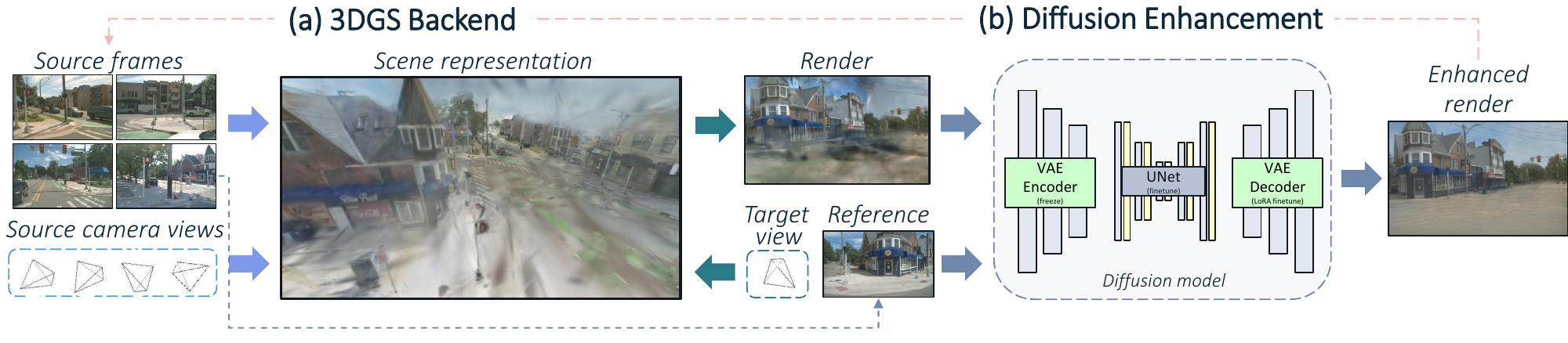}
\vspace{-1em}
\caption{\textbf{Our proposed hybrid NVS pipeline.} First, (a) the source frames and the corresponding views are used to fit a 3DGS model of the scene (blue arrows), that can be used to render frames at the desired target view (green arrows). Then, (b) the rendered view from 3DGS is refined by a dedicated enhancer model, along with a reference source frame (gray arrows). Optionally, the enhanced render can be used as a pseudo source view to further refine the 3DGS reconstruction (red arrow).}
\label{fig:method}
\end{figure*}

The RealADSim-NVS challenge aims to push the state-of-the-art for novel view rendering in street scenes.
Every participating team is provided with 12 video sequences, held out from the Paralane~\cite{paralane} and EUVS~\cite{euvs} datasets, comprising training frames and the corresponding camera viewpoints.
Every sequence also comes with target camera viewpoints, that participants need to generate frames for, whose quality is then measured against groundtruth target images.
More formally, a NVS model is given one or multiple traversals of a driving scene, represented by a set of frames $\{\xi\}_{i=1}^N$, the corresponding camera poses and intrinsic matrices $\{\pi,\ci\}$ and a 3D point cloud $\pc$.
The task is then to generate a set of test frames $\{\yj\}_{j=1}^M$ from the same scene, from cameras $\{\pj,\cj\}$ that correspond to a held-out traversal.
Depending on the nature of the test traversal, three levels of complexity are defined:
\begin{itemize}
\item Level 1: views are shifted by one lane from the training;
\item Level 2: views are shifted by two lanes from the training;
\item Level 3: views are captured from the opposite direction.
\end{itemize}

\noindent The quality of the generated test views is evaluated in terms of PSNR, SSIM and LPIPS~\cite{lpips}.
The three scores are aggregated into a cumulative metric as:
\begin{equation}
\label{eq:challenge_score}
S = 0.4 \times \frac{PSNR}{100} + 0.3 \times SSIM + 0.3 \times (1-LPIPS).
\end{equation}

\section{Hybrid Gaussian Splatting for NVS}
\bgroup
\setlength{\tabcolsep}{1pt}
\renewcommand{\arraystretch}{0.5}
\begin{figure*}[t]
\centering
\resizebox{0.9\textwidth}{!}{
\begin{tabular}{cccc}
Example source view & Target view & 3DGS Backend & Diffusion Enhancer\\
\includegraphics[width=0.25\textwidth]{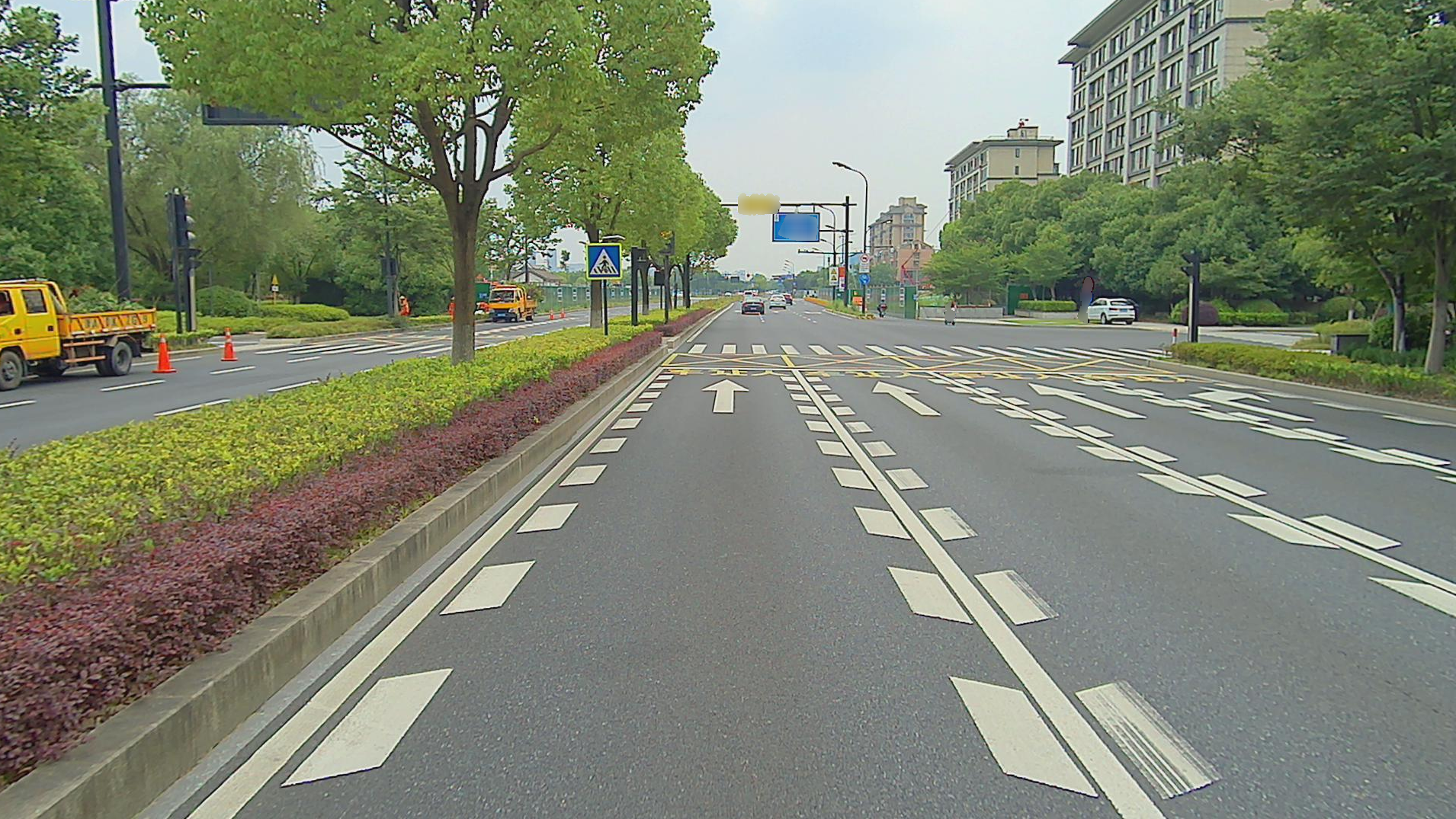}&
\includegraphics[width=0.25\textwidth]{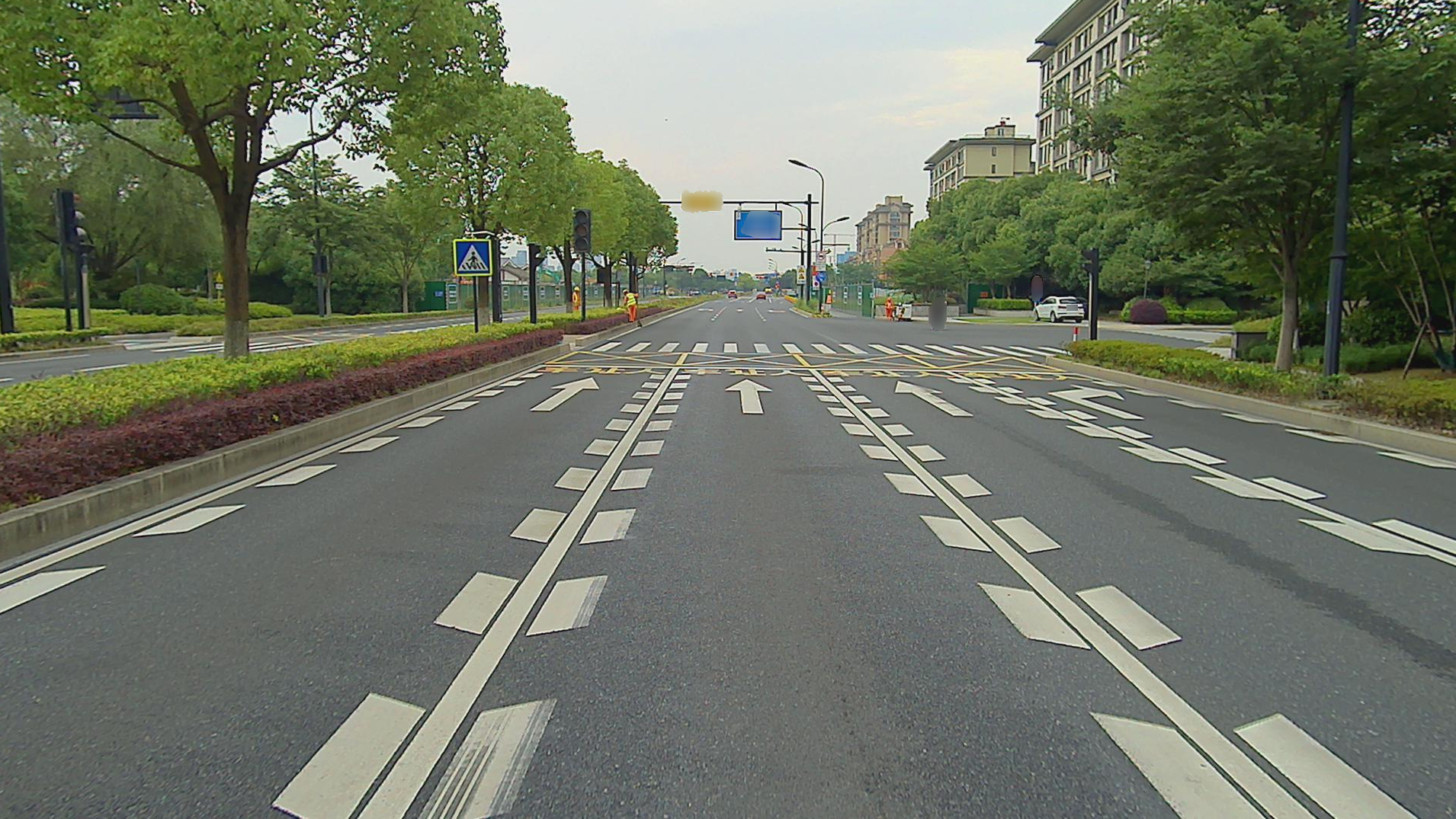}&
\includegraphics[width=0.25\textwidth]{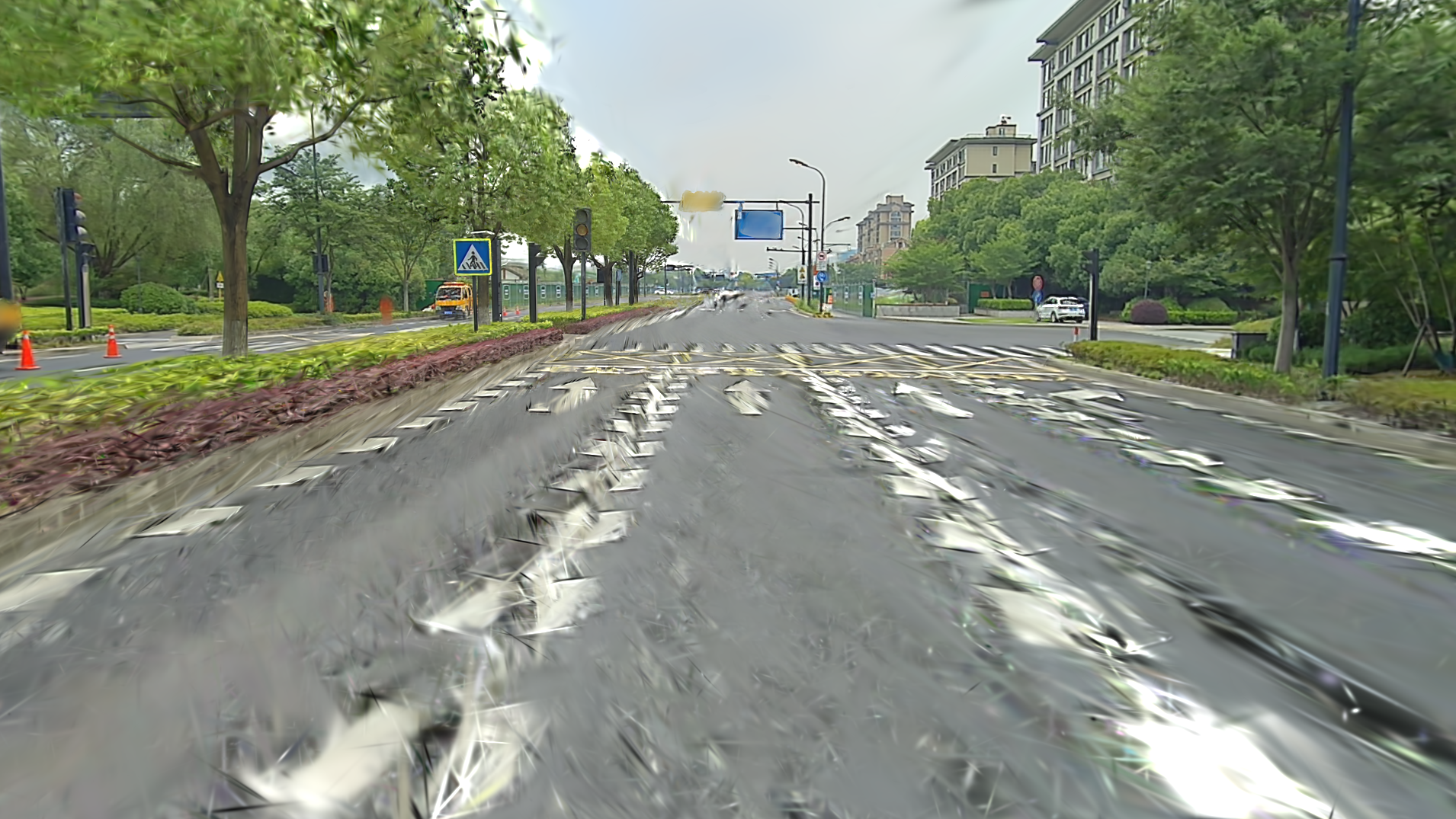}&
\includegraphics[width=0.25\textwidth]{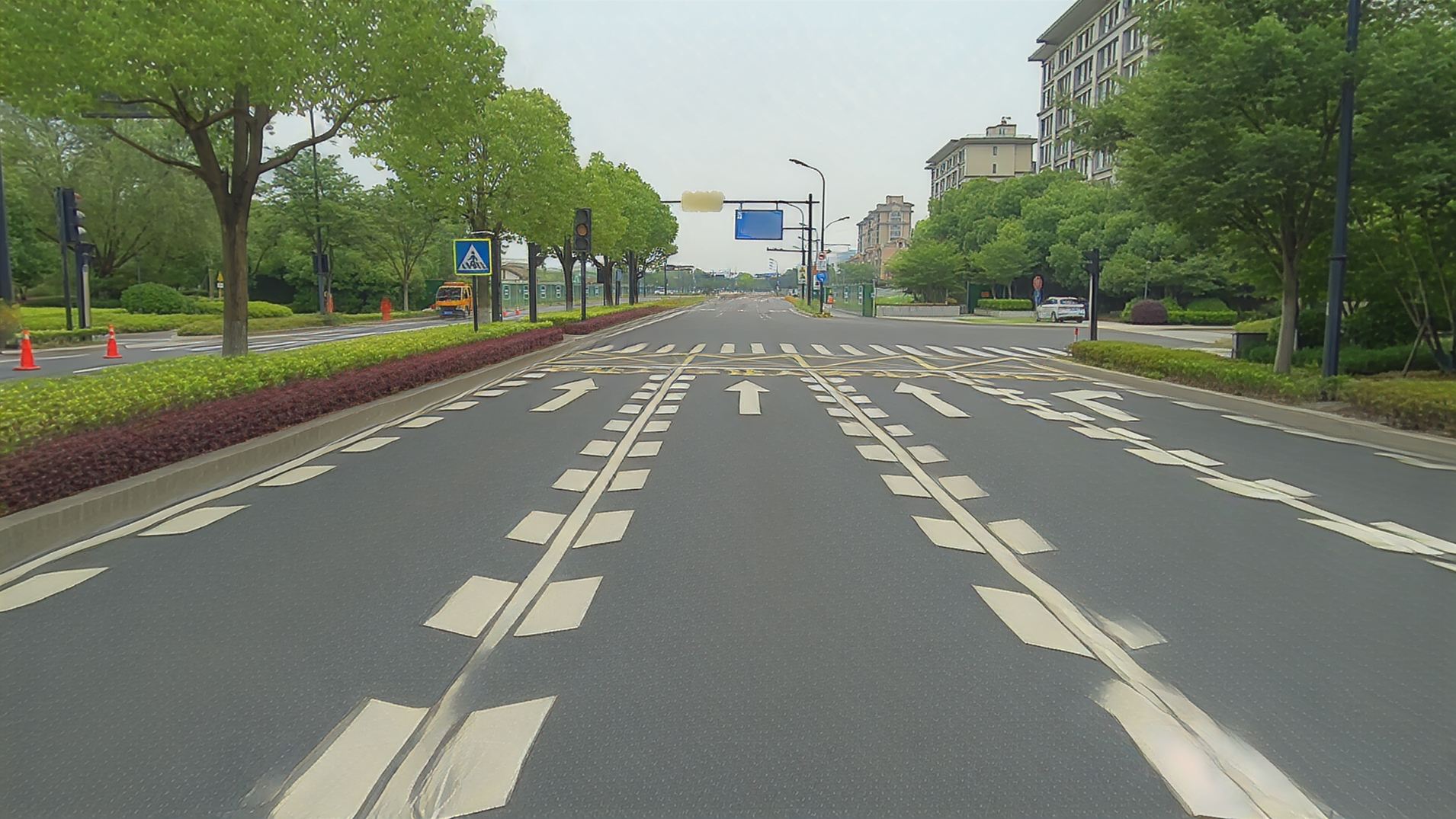}
\\
\includegraphics[width=0.25\textwidth]{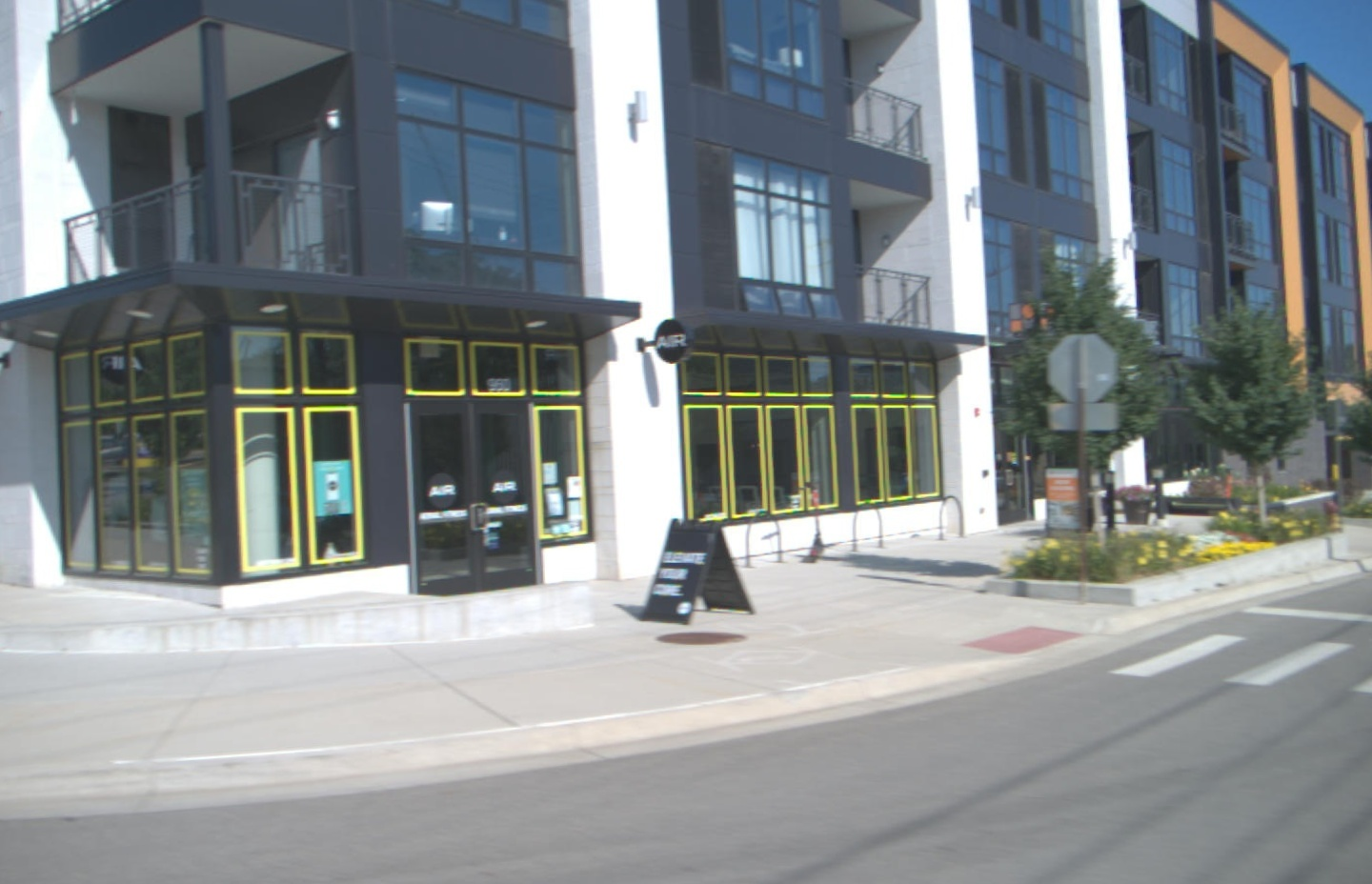}&
\includegraphics[width=0.25\textwidth]{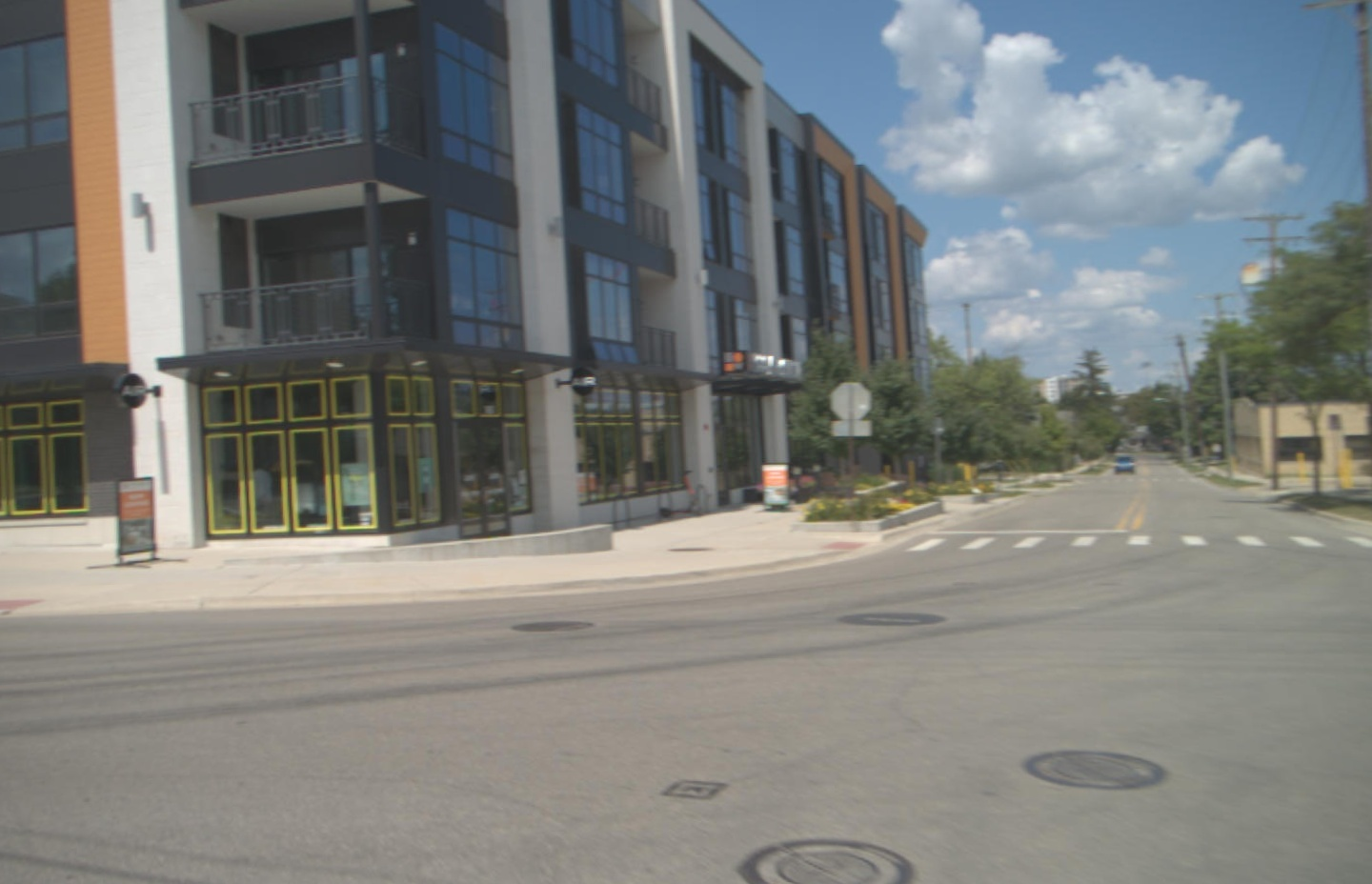}&
\includegraphics[width=0.25\textwidth]{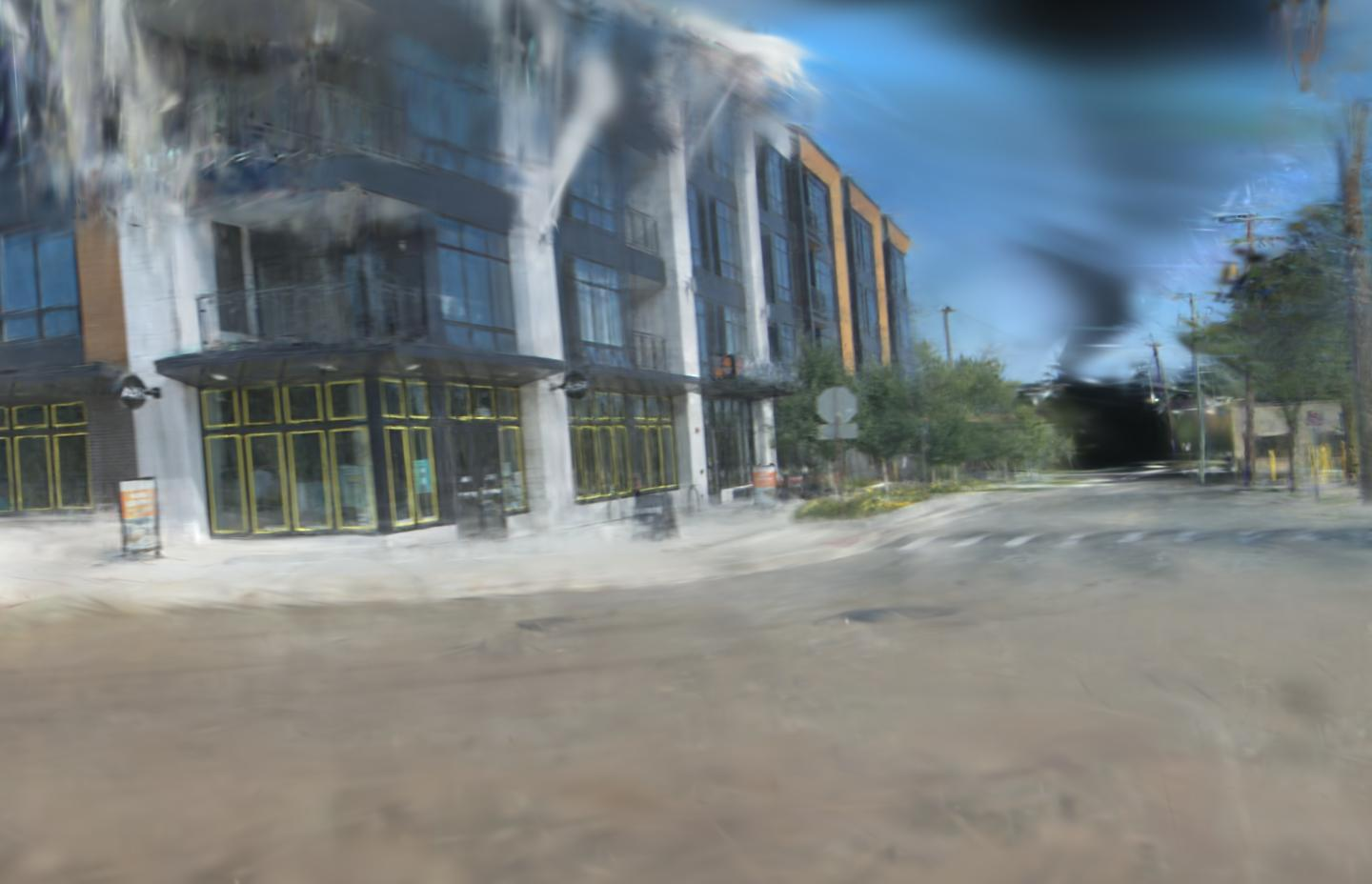}&
\includegraphics[width=0.25\textwidth]{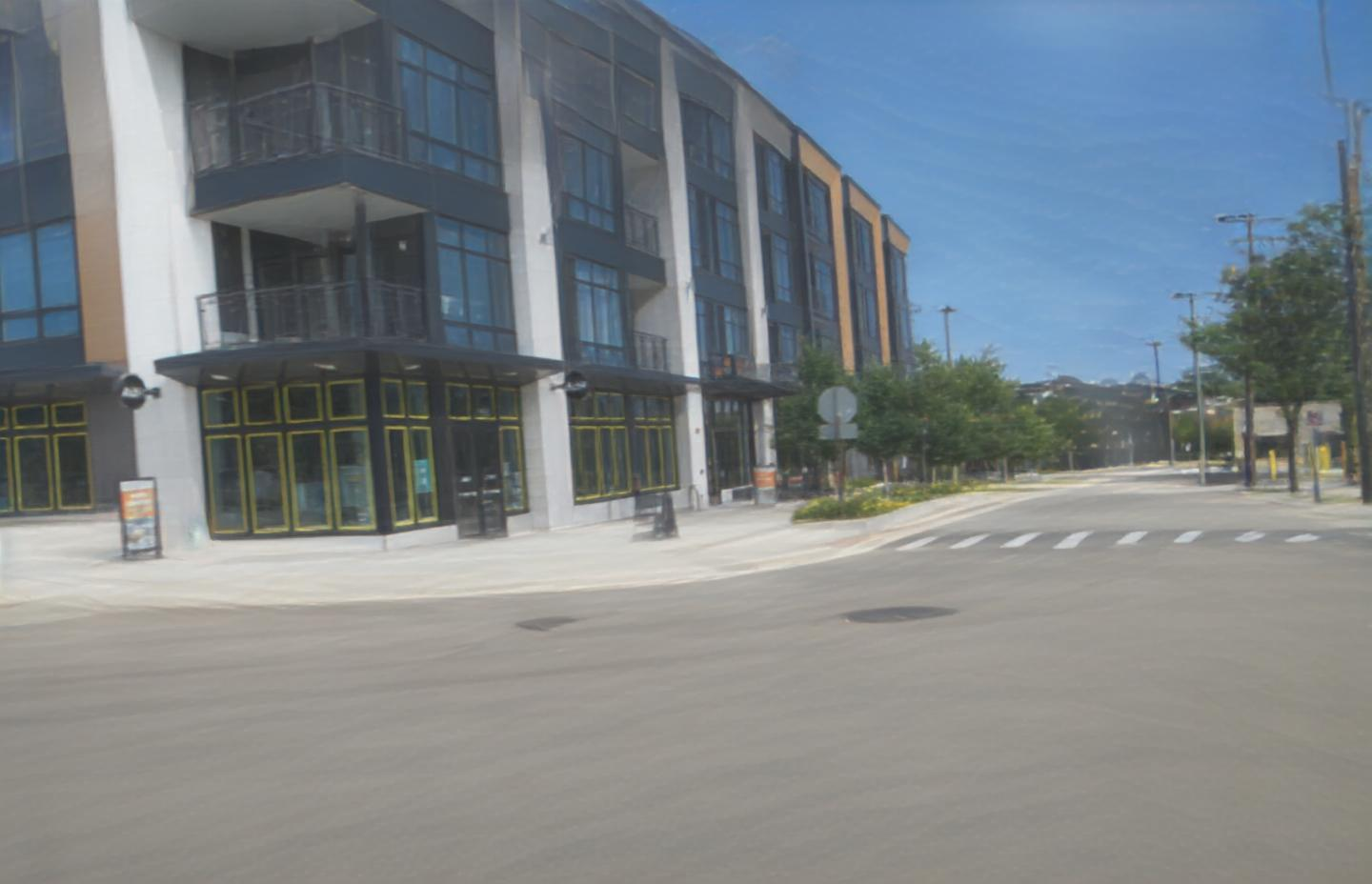}
\\
\includegraphics[width=0.25\textwidth]{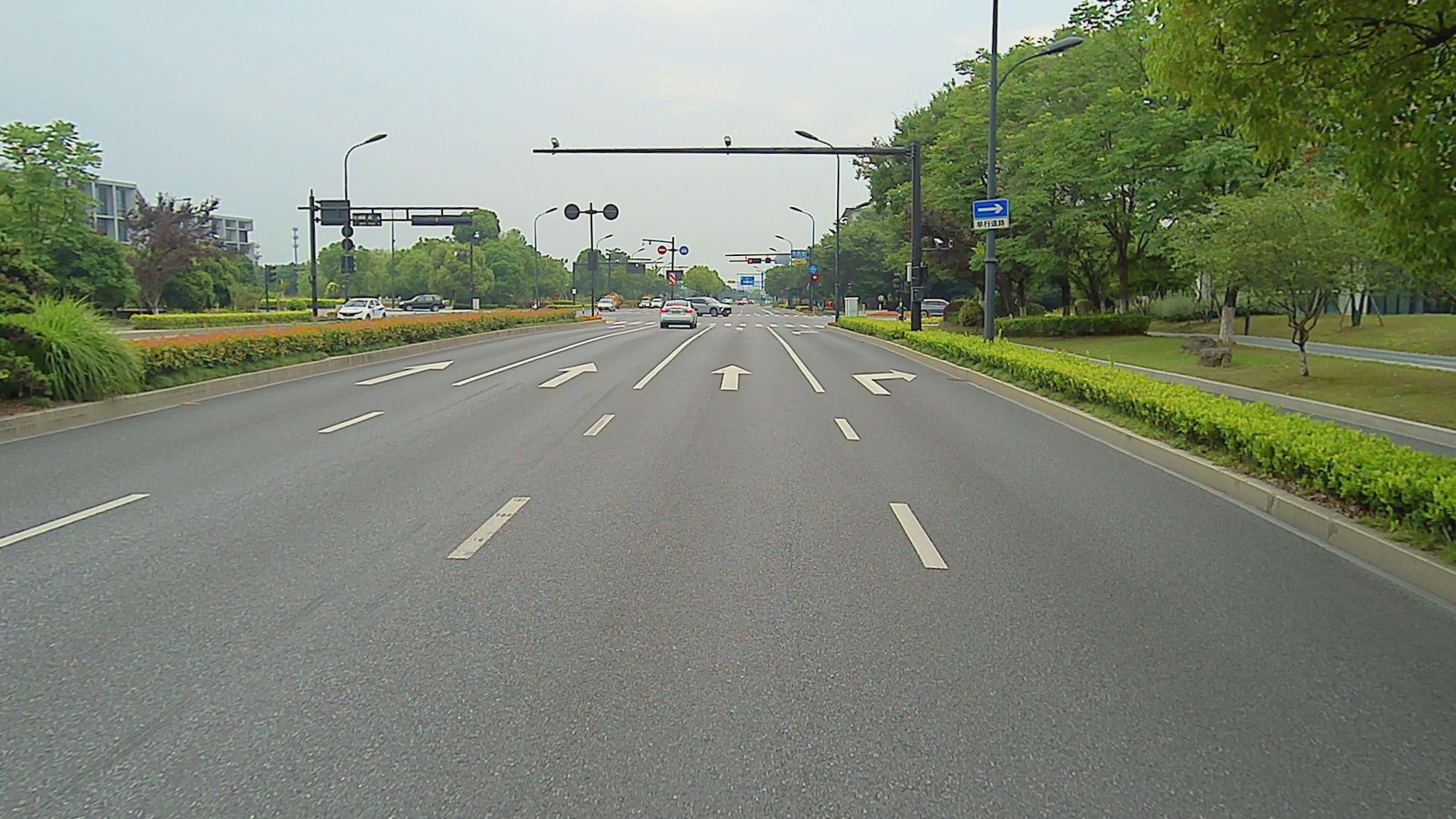}&
\includegraphics[width=0.25\textwidth]{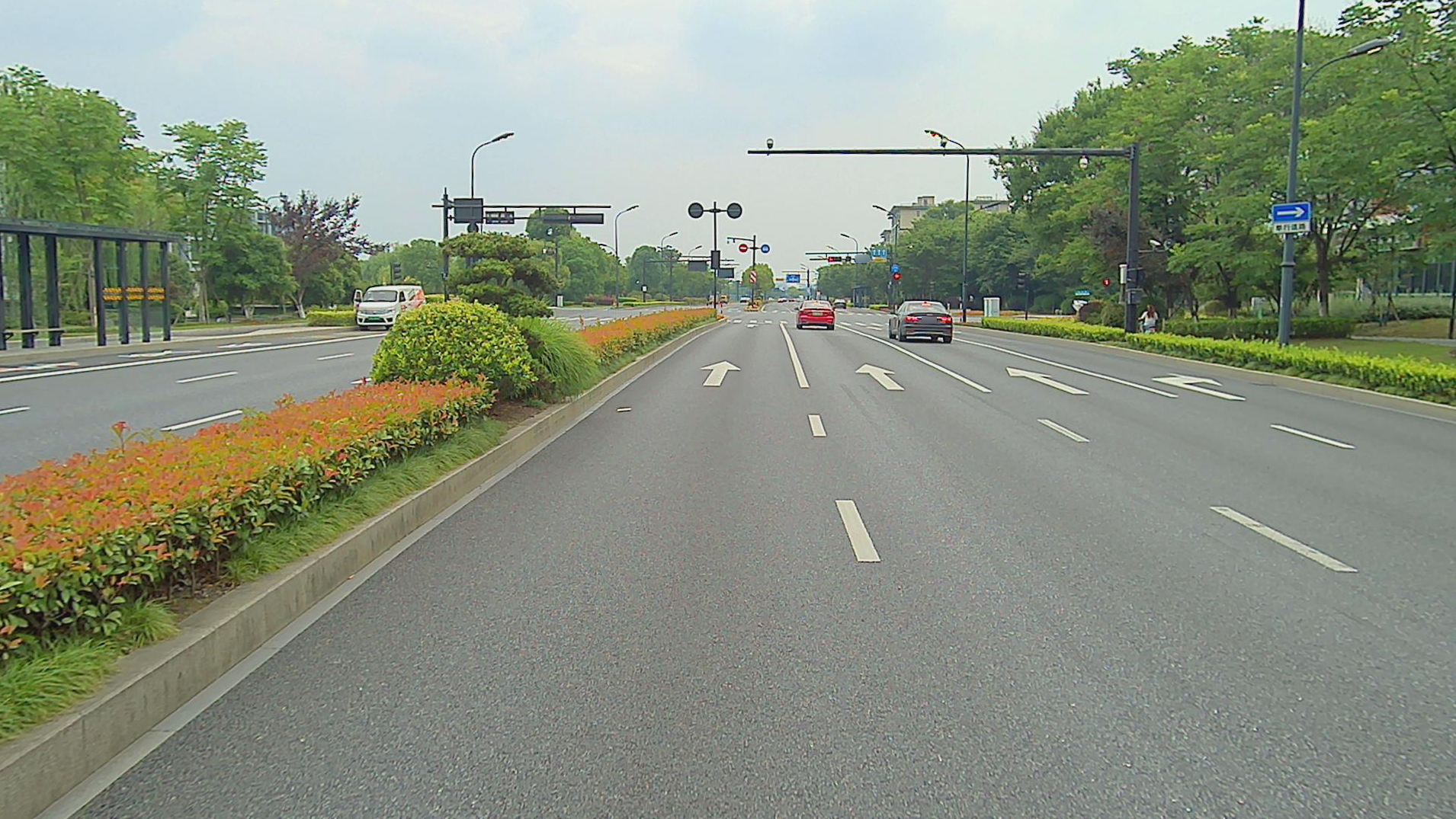}&
\includegraphics[width=0.25\textwidth]{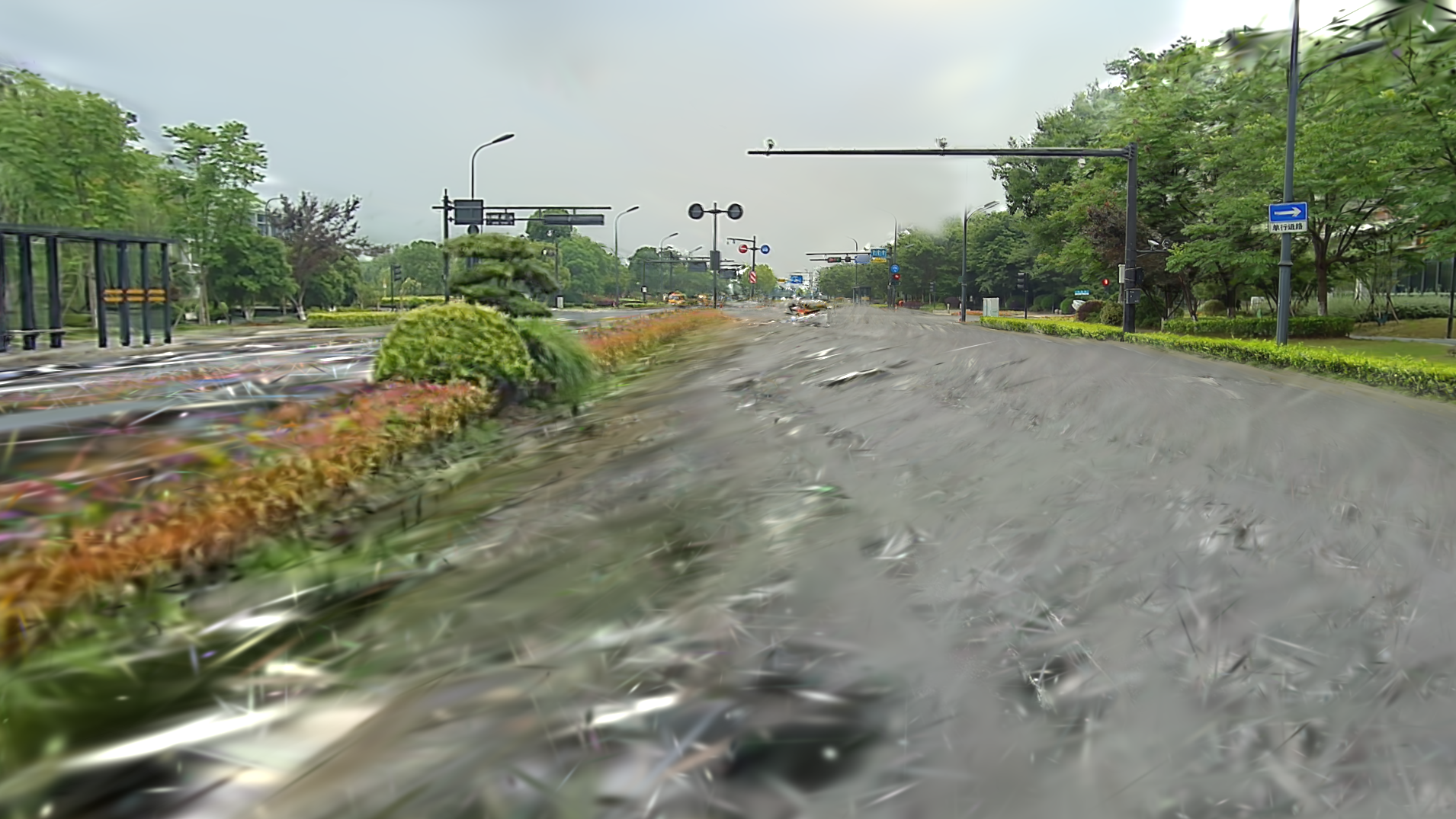}&
\includegraphics[width=0.25\textwidth]{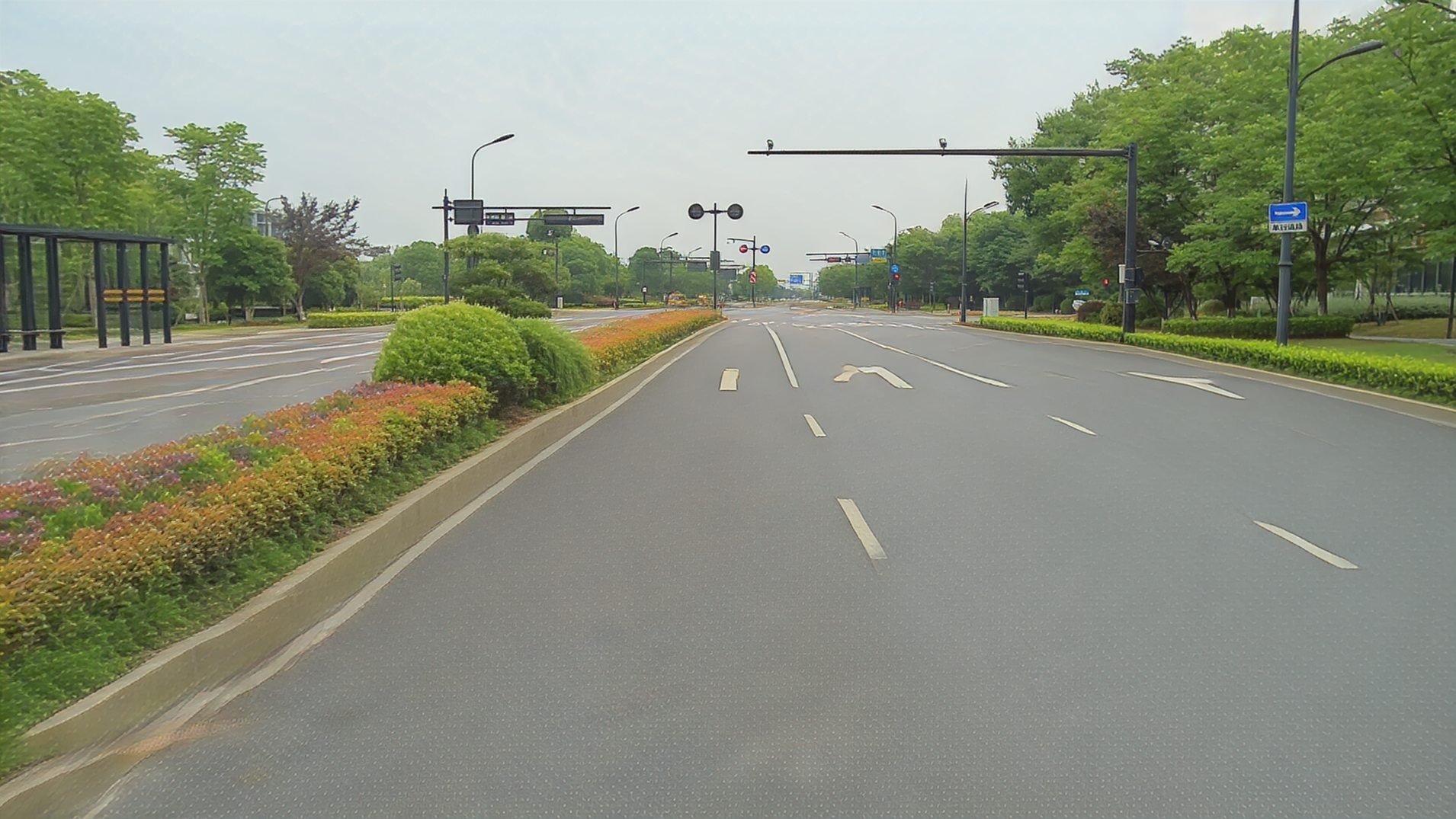}
\\
\includegraphics[width=0.25\textwidth]{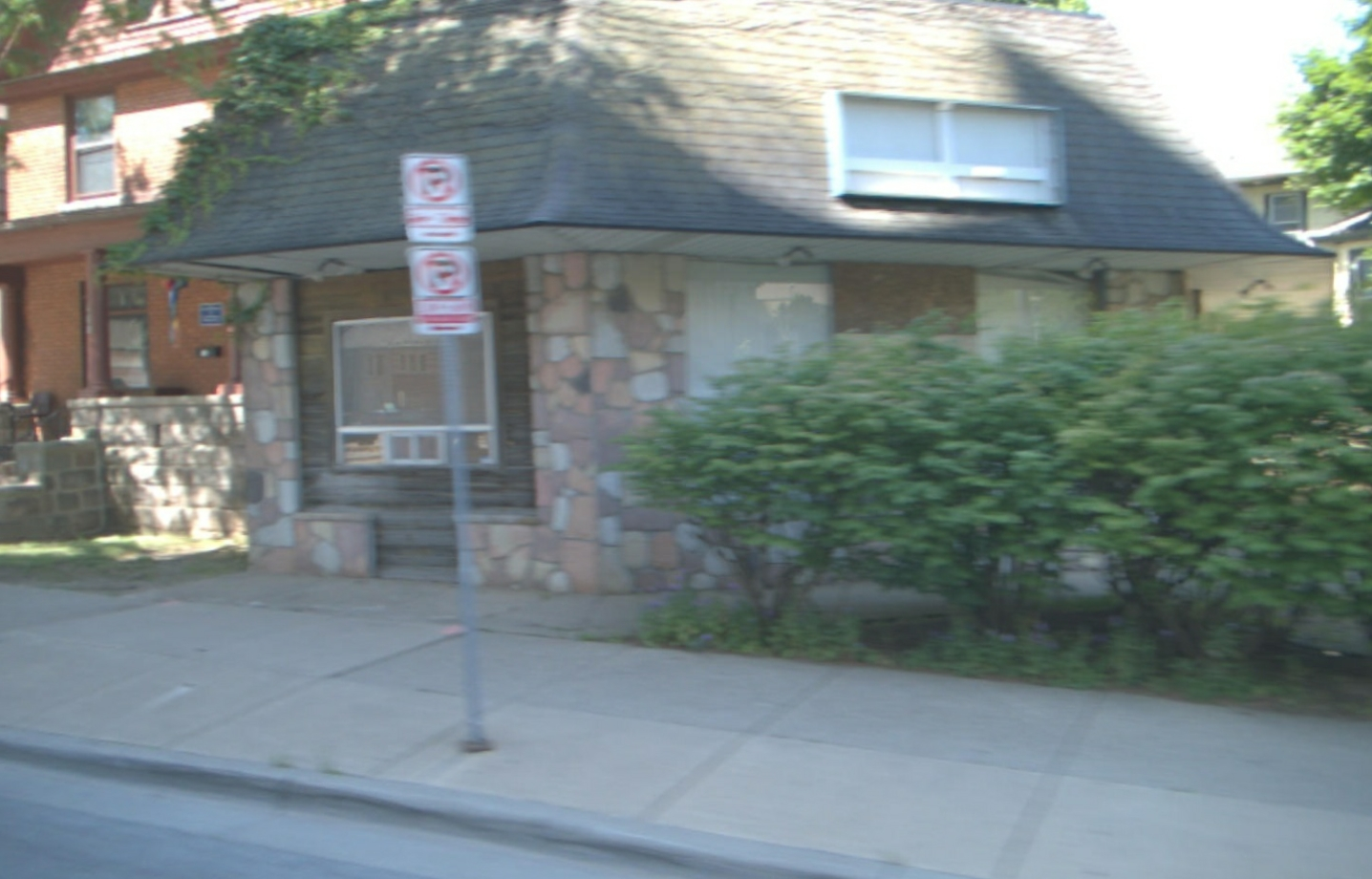}&
\includegraphics[width=0.25\textwidth]{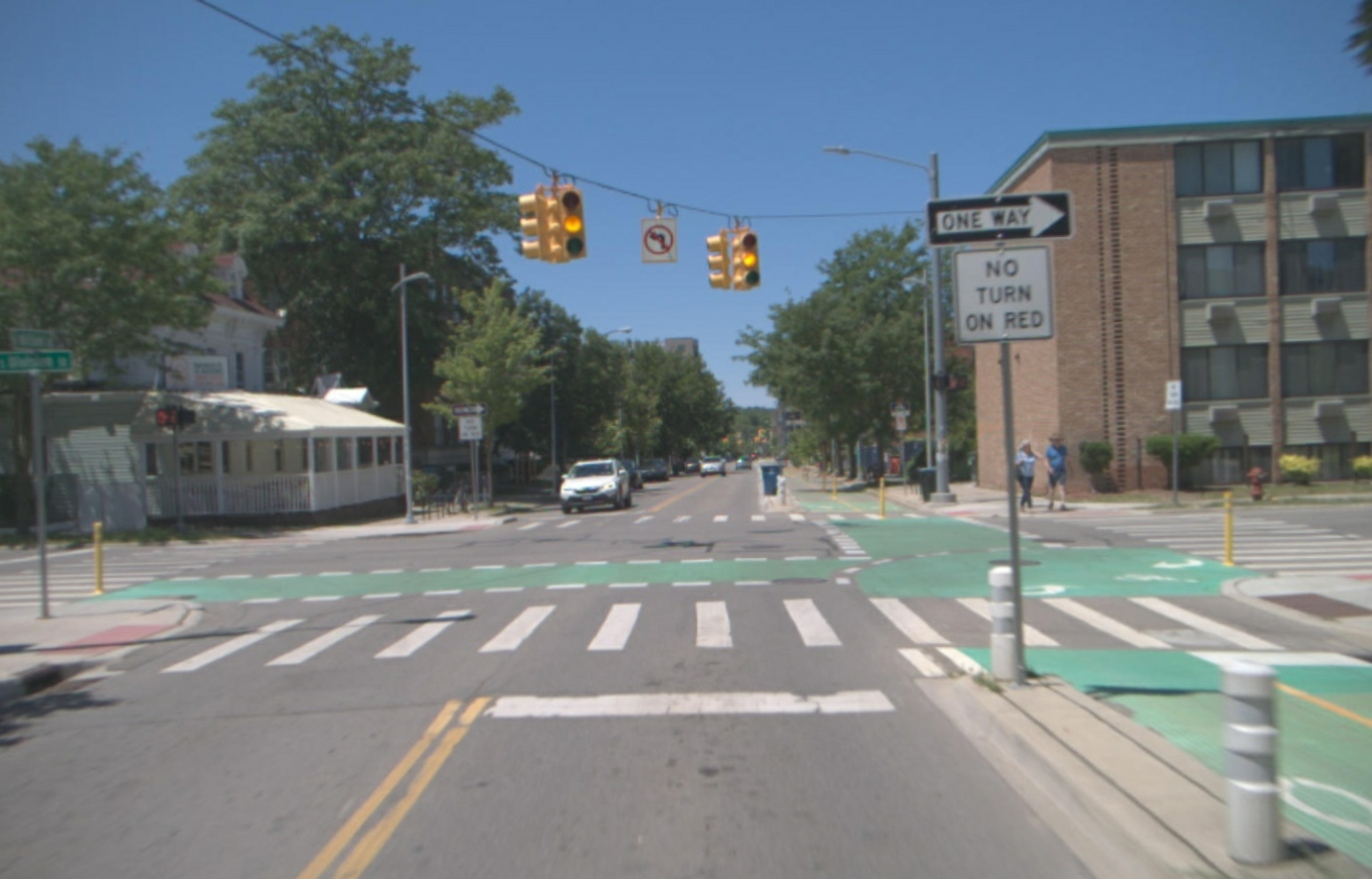}&
\includegraphics[width=0.25\textwidth]{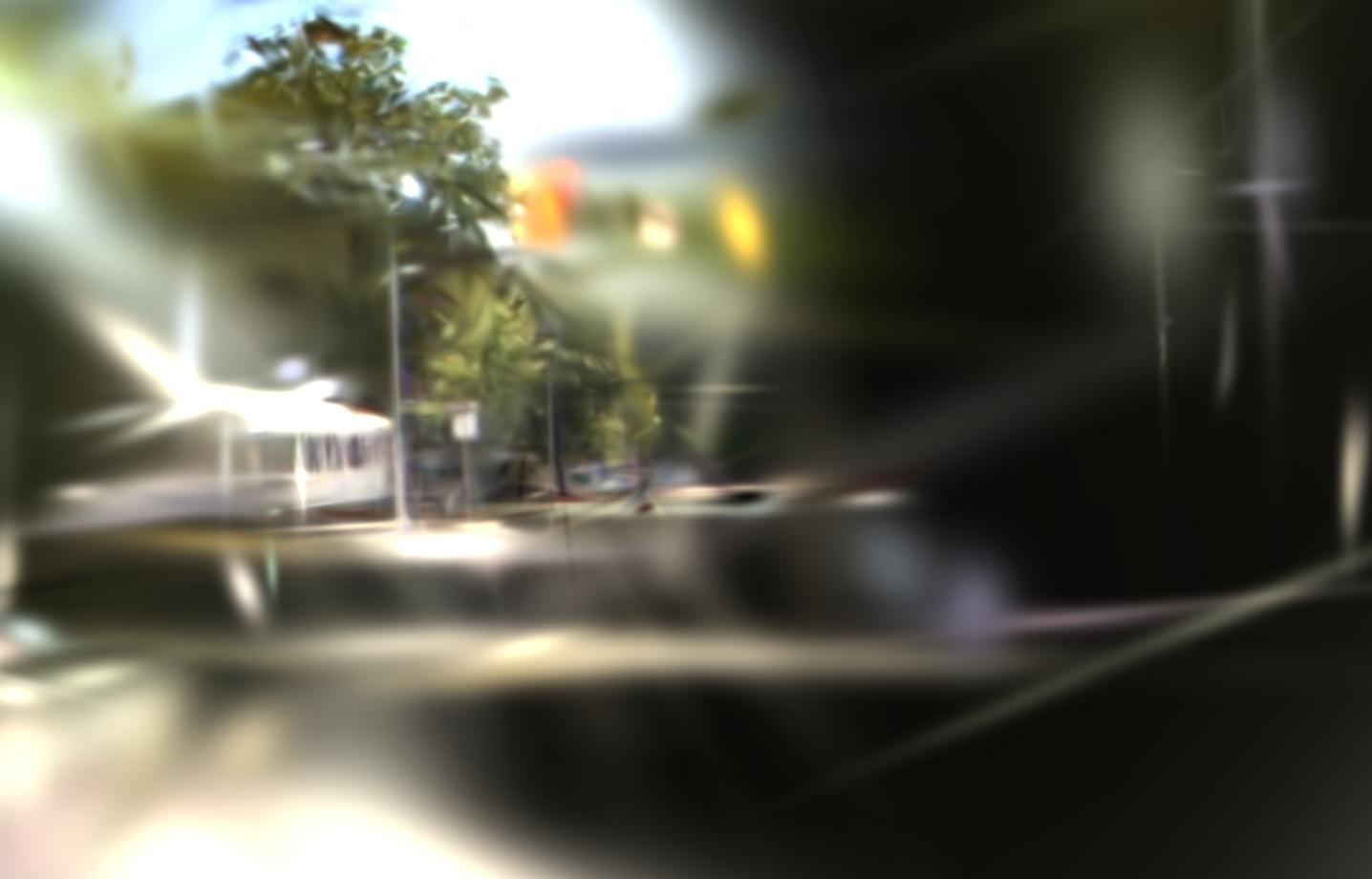}&
\includegraphics[width=0.25\textwidth]{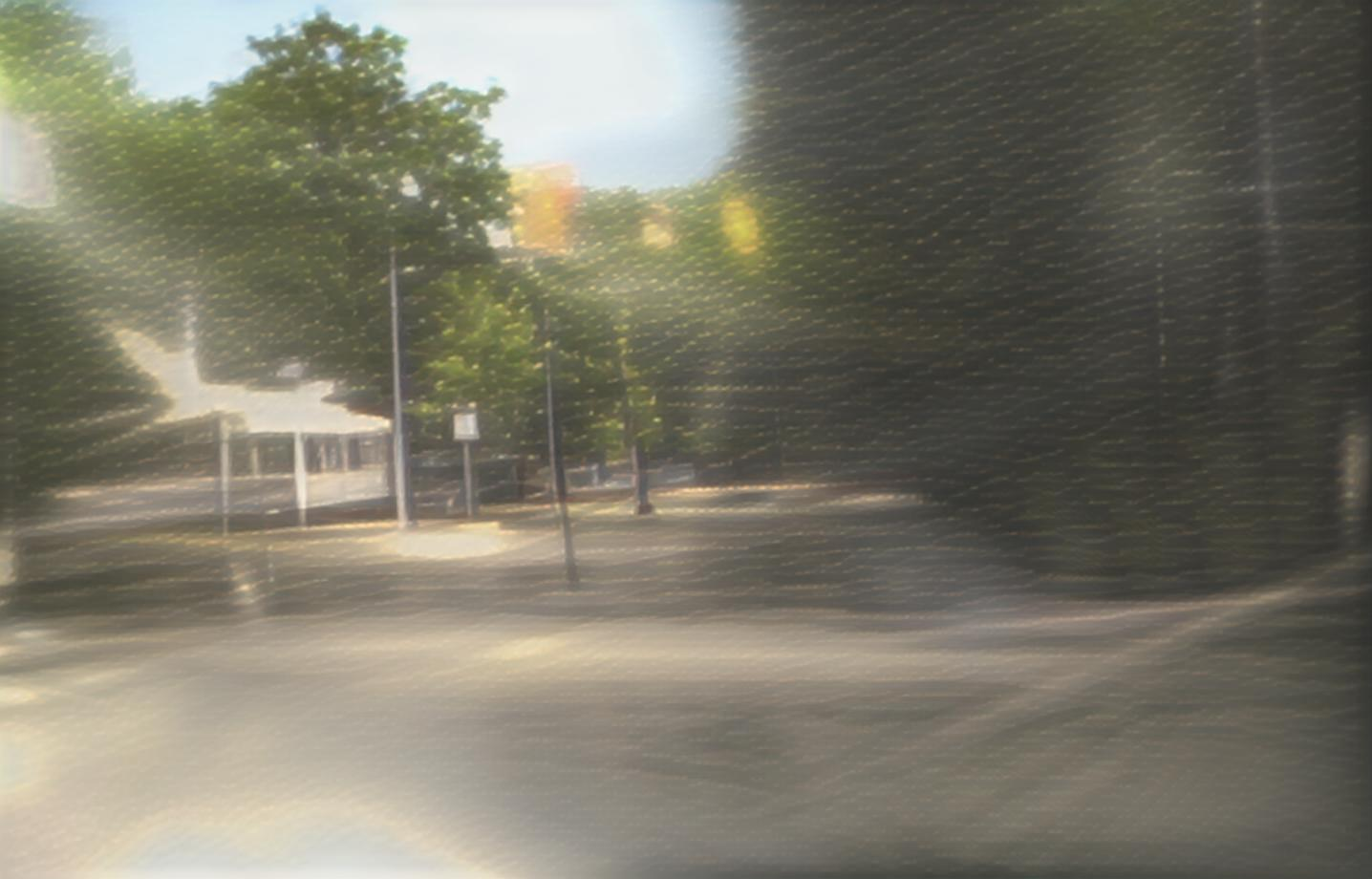}
\end{tabular}}
\caption{\textbf{Qualitative results of our approach.} Fist three rows: our 3DGS backend bootstraps the novel view, creating a frame from the correct viewpoint but featuring novel-view artifacts, especially close to the camera. The enhancer model then polishes the frame yielding the final frame. Last row: a failure case of our approach. Whenever the distance between source and target viewpoints is too high, the 3DGS backend fails completely, and the enhancer cannot recover a meaningful frame.}
\label{fig:qualitative}
\end{figure*}
\egroup

Our proposed hybrid pipeline is composed of two stages, as depicted in \cref{fig:method}. First, the geometry of the scene is reconstructed with 3D Gaussian Splatting, and this representation is used to bootstrap the frames as captured from the test views. Then, the rendering artifacts are corrected with a single-step diffusion enhancer.
\subsection{Gaussian Splatting Backend}
\label{sec:3dgs}
3D Gaussian Splatting (3DGS,~\cite{3dgs}) allows to compute an explicit representation of a scene by means of a set of Gaussian primitives.
At a high level, the 3DGS framework defines a fitting function like
\begin{equation}
    \gsmodel \leftarrow Fit(\{\xi,\pi,\ci\},\pc),
\end{equation}
where $\gsmodel$ is a scene reconstruction composed of millions of Gaussians, each parametrized by its position, rotation, scale, view-dependent color and opacity.
Once fitted, $\gsmodel$ is used to render the scene from an arbitrary novel viewpoint:
\begin{equation}
    \ygsj = Render(\gsmodel,\pj,\cj).
\end{equation}
We refer the interested reader to~\cite{3dgs} for more details on how to train and render from 3DGS models.

Within this 3DGS framework, we experimented with several variants: (i) different primitive types (3D Gaussians~\cite{3dgs} vs. Triangles~\cite{Held2025Triangle}), (ii) different densification strategies (default~\cite{3dgs} vs. MCMC-based~\cite{NEURIPS2024_93be245f}), (iii) regularisation with geometric priors (using monocular depth and surface normals~\cite{vegs, turkulainen2024dnsplatter}), and (iv) separate treatment of sky~\cite{Held2025Triangle, ye2024gaustudio} and ground~\cite{hugsim}. 
Ultimately, we found that all these alternatives do not outperform the standard 3DGS implementation, provided that the model was initialized carefully.
Indeed, during the fitting stage, the point cloud $\pc$ is typically used to initialize the 3D locations of the gaussians primitives prior to optimization.
Whereas the challenge sequences include pre-computed point clouds, early inspections revealed them to be very sparse (e.g. on the order of $~1e^{3}$ points for the Paralane sequences). 
We therefore re-estimate them given the images and the camera information, to benefit from a higher number of 3D points.
Using the COLMAP open source library~\cite{colmap1,colmap2}, we first estimate a sparse point cloud given known camera poses and intrinsics. 
This involves feature extraction, feature matching, and point triangulation and results in around $~1e^{5}$ per sequence. 
We then additionally apply the standard dense reconstruction pipeline, consisting of image undistortion, depth and normal map estimation, as well as a final step fusing latter maps to obtain a point cloud. 
This procedure results in several millions of points per sequence, and significantly improves the quality of the resulting renders.
\subsection{Diffusion Enhancement}
Whereas the 3DGS framework easily allows for rendering the scene from an arbitrary camera pose, the rendered frames can exhibit significant artifacts, that get more severe the more the viewpoint is far from the training ones.
For this reason, we rely on an enhancer to compensate for such artifacts.
We derive its architecture from SD-Turbo~\cite{sdturbo,difix}, and we finetune it for the task of the removal of 3DGS rendering artifacts.
Specifically, we postprocess the 3DGS render $\ygsj$ with the enhancer model $\difix$, as
\begin{equation}
    \yhatj = \difix(\ygsj,\xref),
\end{equation}
where $\yhatj$ is the enhanced frame for the target view $\pj,\cj$ and $\xref$ is a clean reference frame, sampled from the source frames by minimizing a camera pose distance to the target (\ie $r=\argmin_i d(\pi,\pj))$, with $d$ a distance measure between camera poses\footnote{We use a weighted sum of translation distance and quaternion rotation distance, with $\lambda_{tr}=1$ and $\lambda_{rot}=10$.}).
We refer to~\cite{sdturbo,difix} for further details about the diffusion enhancement architecture.
\paragraph{Training data curation.}
To properly train the enhancer model, we need a large dataset of frames exhibiting 3DGS rendering artifacts, each paired with its corresponding clean counterpart.
To this aim, we exploit the public sequences of the Paralane~\cite{paralane} and EUVS~\cite{euvs} datasets.
Paralane features 25 public sequences featuring three clips each, where each clip corresponds to a traversal in the same three lane street.
Here, we fit our 3DGS implementation on every clip, and use such models to render the other ones, resulting in 150 clips with rendering artifacts coming from one-lane or two-lane shifts.
Concerning EUVS, we rely on the 79 public scenes, each featuring a predefined split between training and testing traversals: we conform to such splits to render corrupted frames with 3DGS.
Overall, the mixture of the two datasets yields 36,231 training pairs of corrupted and clean frames. 
Importantly, as the source and target frames that we use to fit and render from 3DGS belong to different traversals of the same scene, they often contain different dynamic objects (\eg cars, pedestrians). 
For this reason, when fitting 3DGS, we mask out dynamic objects from the loss, following segmentation masks obtained by Grounded Sam 2~\cite{grounded_sam}.
As a result of this procedure, the frames frames that we render lack these objects and depict empty scenes.

\paragraph{Training loss.}
We train our model by minimizing a weighted sum of MSE, SSIM and LPIPS objectives between the enhanced frames and the groundtruth clean frames.
Similarily to the case of 3DGS fitting and rendering described above, the input training frames (and in turn the corresponding enhanced prediction) depict empty street scenes, whereas the corresponding groundtruth might feature some dynamic objects such as cars and pedestrians.
To avoid noise in the training signal, we mask out the latters from both the prediction and the groundtruth prior to computing each loss component, again using~\cite{grounded_sam}.

\paragraph{Training resolution.}
We experimented with several operating resolutions for the enhancer, trading off quality and training memory requirements.
We found that running the enhancer at $720\times1280$ pixels yielded the best results.

\paragraph{Iterative 3DGS reconstruction and rendering.}
At test time, we can optionally use the enhanced renders to improve the 3DGS reconstruction itself, simply by adding them to the source views and further optimizing the geometry of the scene.
Similarily to~\cite{difix}, we define a $K$-steps pose interpolation between the each target view and the closest source camera.
Before generating frames for the target views, we render the interpolated ones, enhance them, and use them as new pseudo source views to further optimize the 3DGS representation.
We carry out this $K$-steps procedure iteratively, until all interpolated poses are added to the source views: this strategy improves gaussian splatting representation that we ultimately use to render the target view.

\begin{figure}[t!]
\centering
\bgroup
\setlength{\tabcolsep}{1pt}
\resizebox{\columnwidth}{!}{
\begin{tabular}{ccc}
Challenge & Sparse & Dense \\
\includegraphics[width=0.4\columnwidth]{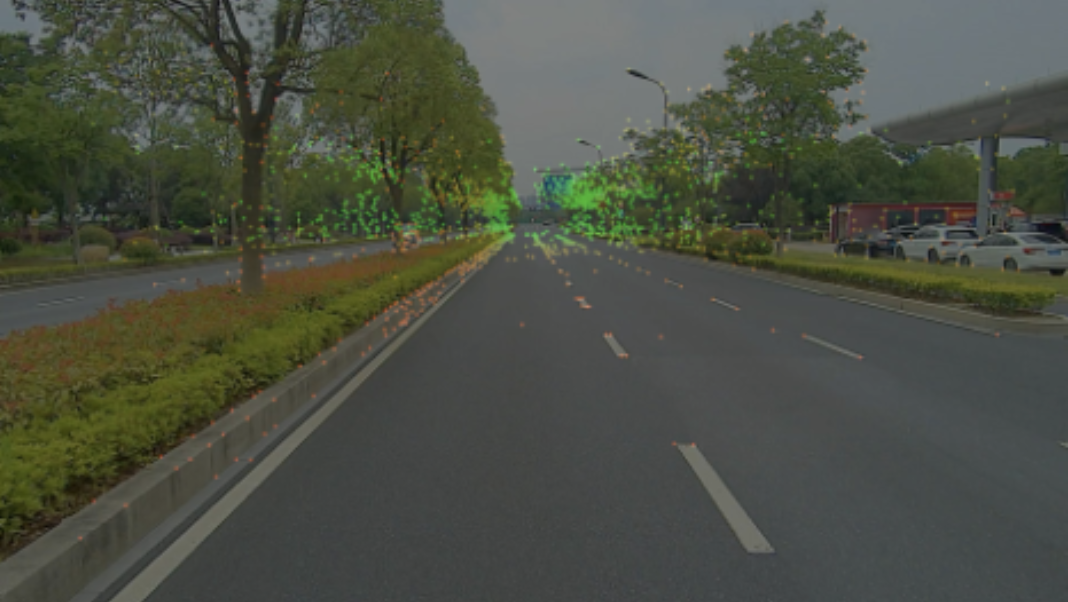} &
\includegraphics[width=0.4\columnwidth]{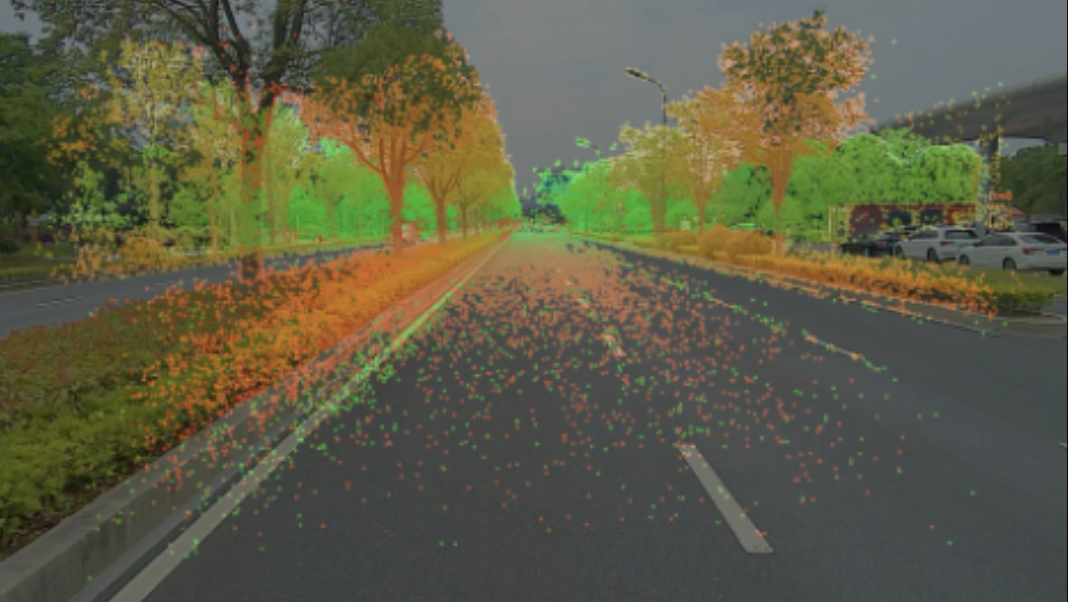} &
\includegraphics[width=0.4\columnwidth]{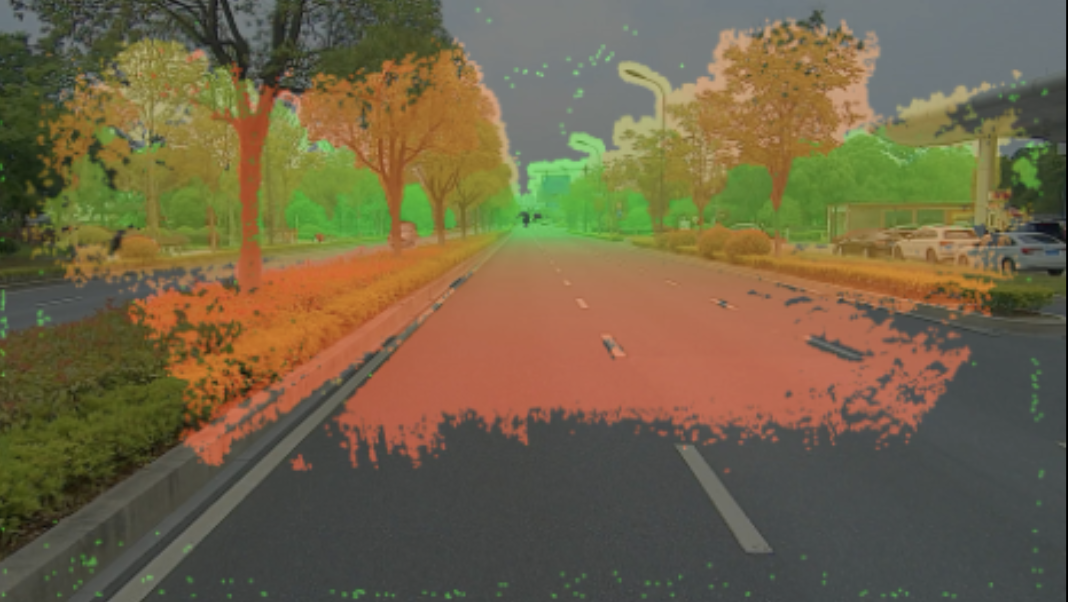}
\end{tabular}}
\egroup
\resizebox{\columnwidth}{!}{
\begin{tabular}{lccccc}
\toprule
& \# pts. & PSNR & SSIM & LPIPS & Score\\
\midrule
Challenge & $\sim$ 30K & 21.03/14.89 & .695/.479 & .326/.614 & .495/.319 \\
Sparse    & $\sim$ 180K & 23.19/15.41 & .769/\textbf{.483} & .231/.545 & .554/.343 \\
Dense     & $\sim$ 5.6M & \textbf{23.93}/\textbf{15.54} & \textbf{.811}/.465 & \textbf{.175}/\textbf{.518} & \textbf{.586}/\textbf{.346} \\
\bottomrule
\end{tabular}}
\caption{\textbf{Impact of point cloud initialization on the 3DGS performance.} By initializing the gaussian locations with more points, the overall NVS performance improves. Numbers reported on the challenge sequences, in the format `source views / target views'.}
\label{tab:initialization}
\vspace{-0.5em}
\end{figure}

\begin{table}[b]
\centering
\resizebox{\columnwidth}{!}{
\begin{tabular}{lcccc}
\toprule
& PSNR & SSIM & LPIPS & Score\\
\midrule
\rowcolor{Gray!30}
Pretrained ($512\times 1024$) & 16.615 & 0.467 & 0.278 & 0.423\\
\quad + FT on curated data  & 18.432  & 0.526  & 0.255 & 0.455\\
\quad + HR training ($720\times 1280$) & 18.463  & 0.522  & \textbf{0.229}  & \textbf{0.462}\\
\quad + Masked loss & \textbf{18.505}  & \textbf{0.523}  & 0.230  & \textbf{0.462}\\
\bottomrule
\end{tabular}}
\caption{\textbf{Impact of enhancer finetuning.} Our data curation strategy proves beneficial for the enhancer. Other benefits come from high resolution training and removing the dynamic objects from the loss computation. Numbers reported on the devset.}
\label{tab:enhancer}
\end{table}

\bgroup
\setlength{\tabcolsep}{1pt}
\renewcommand{\arraystretch}{0.5}
\begin{figure}[t]
\centering
\resizebox{0.95\columnwidth}{!}{
\begin{tabular}{cc}
Enhanced Render & Iter. 3DGS refinement ($K=3$)\\
\includegraphics[width=0.6\columnwidth]{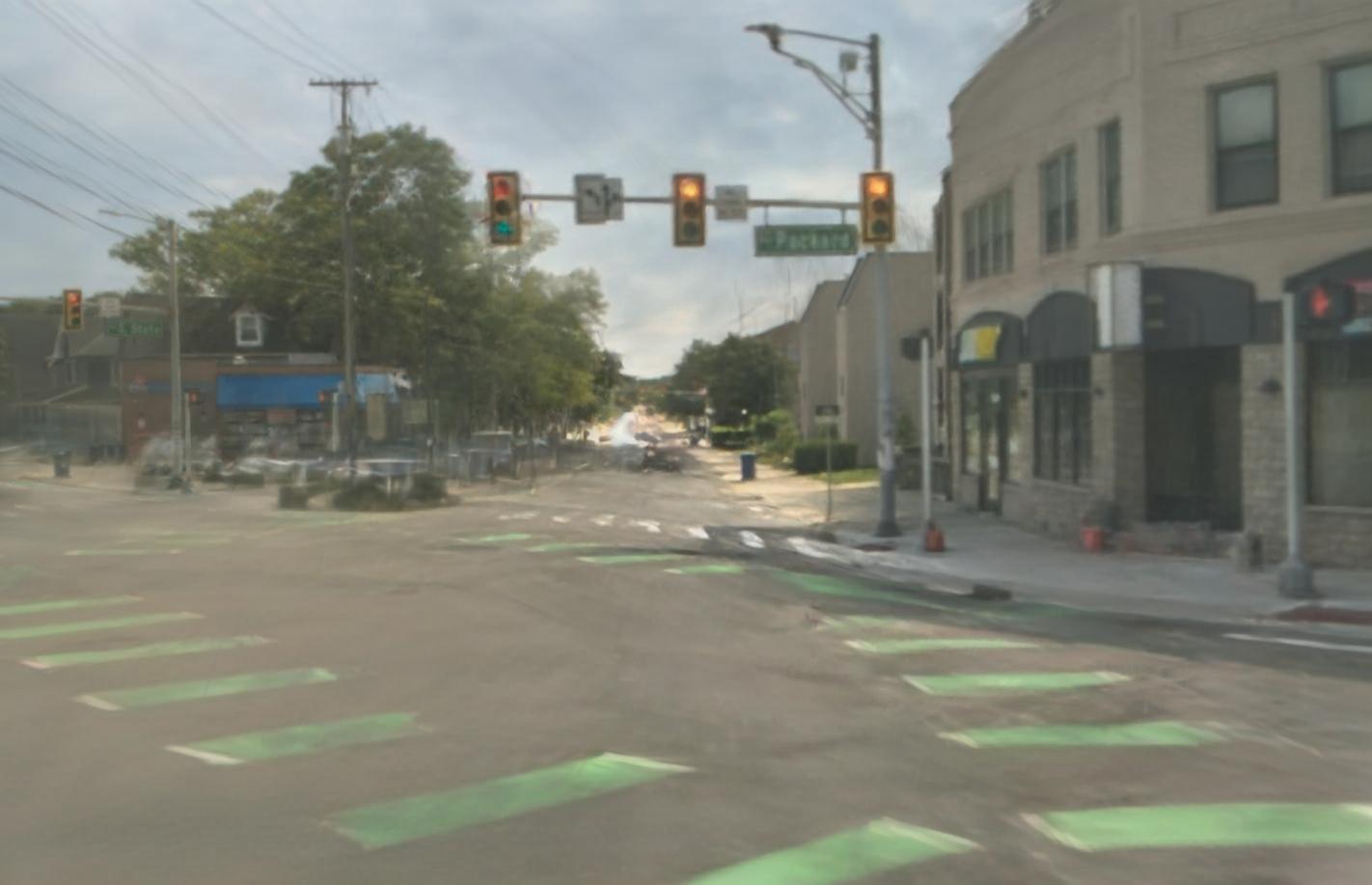}&
\includegraphics[width=0.6\columnwidth]{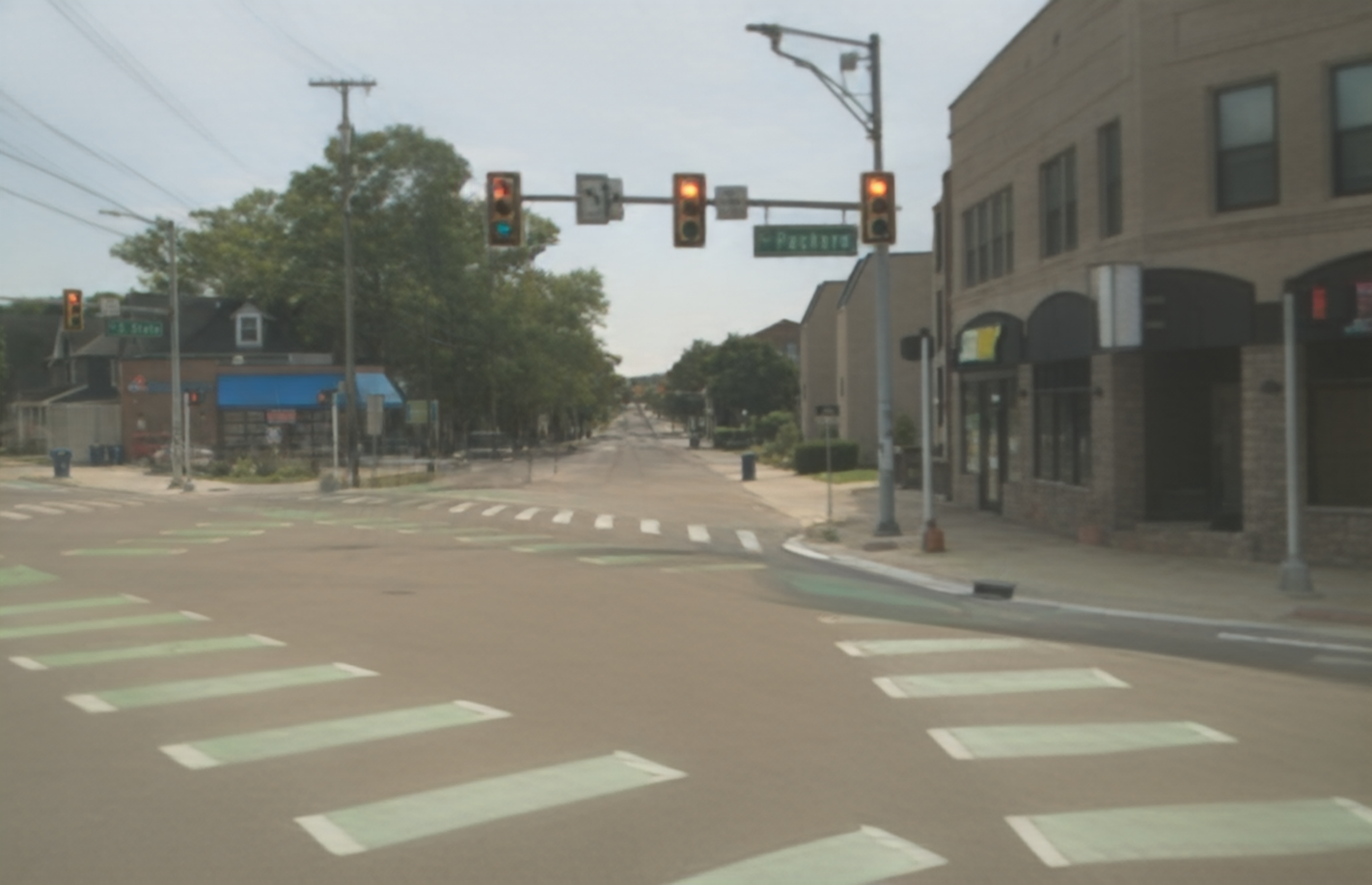}
\\
\includegraphics[width=0.6\columnwidth]{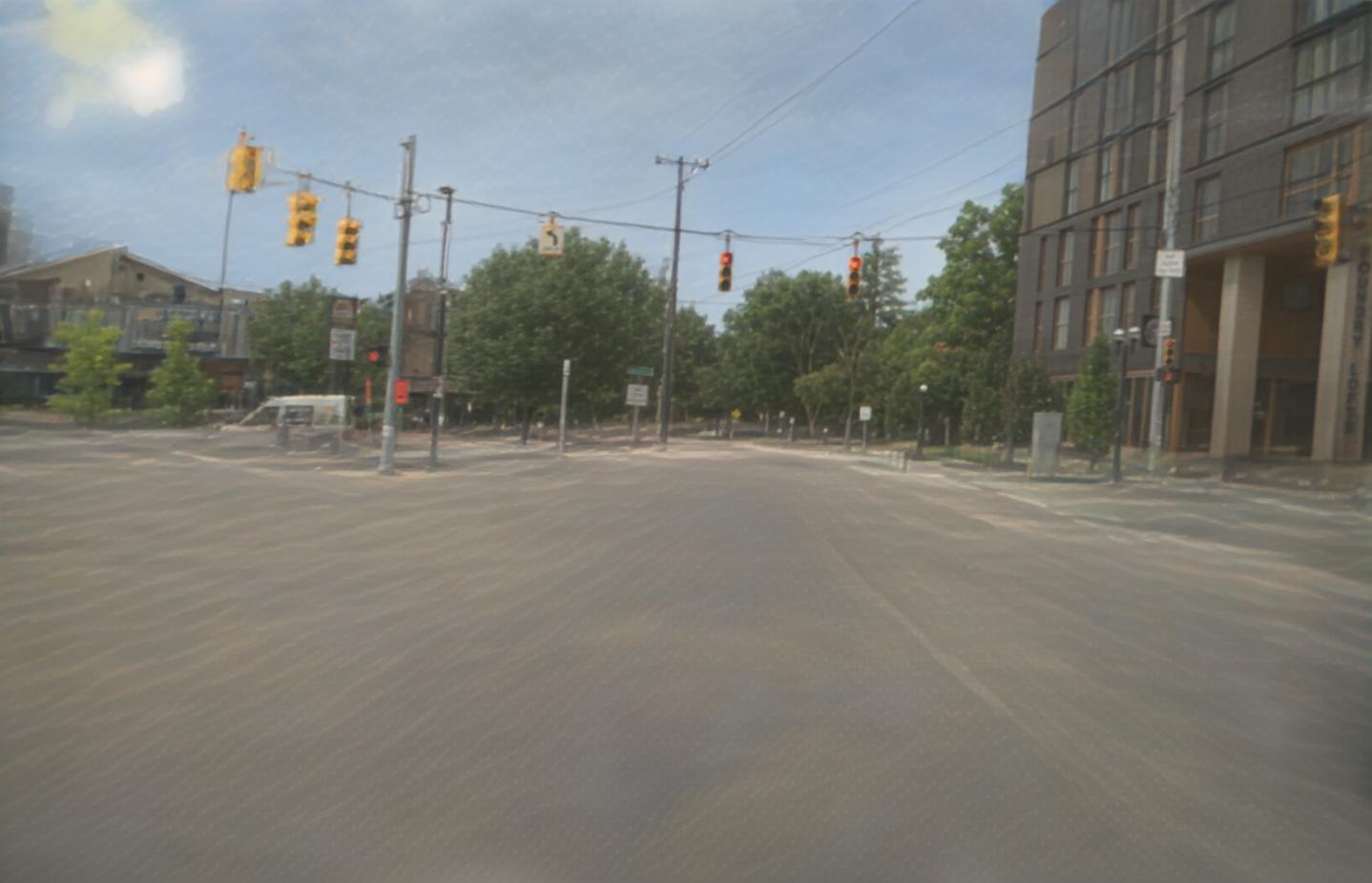}&
\includegraphics[width=0.6\columnwidth]{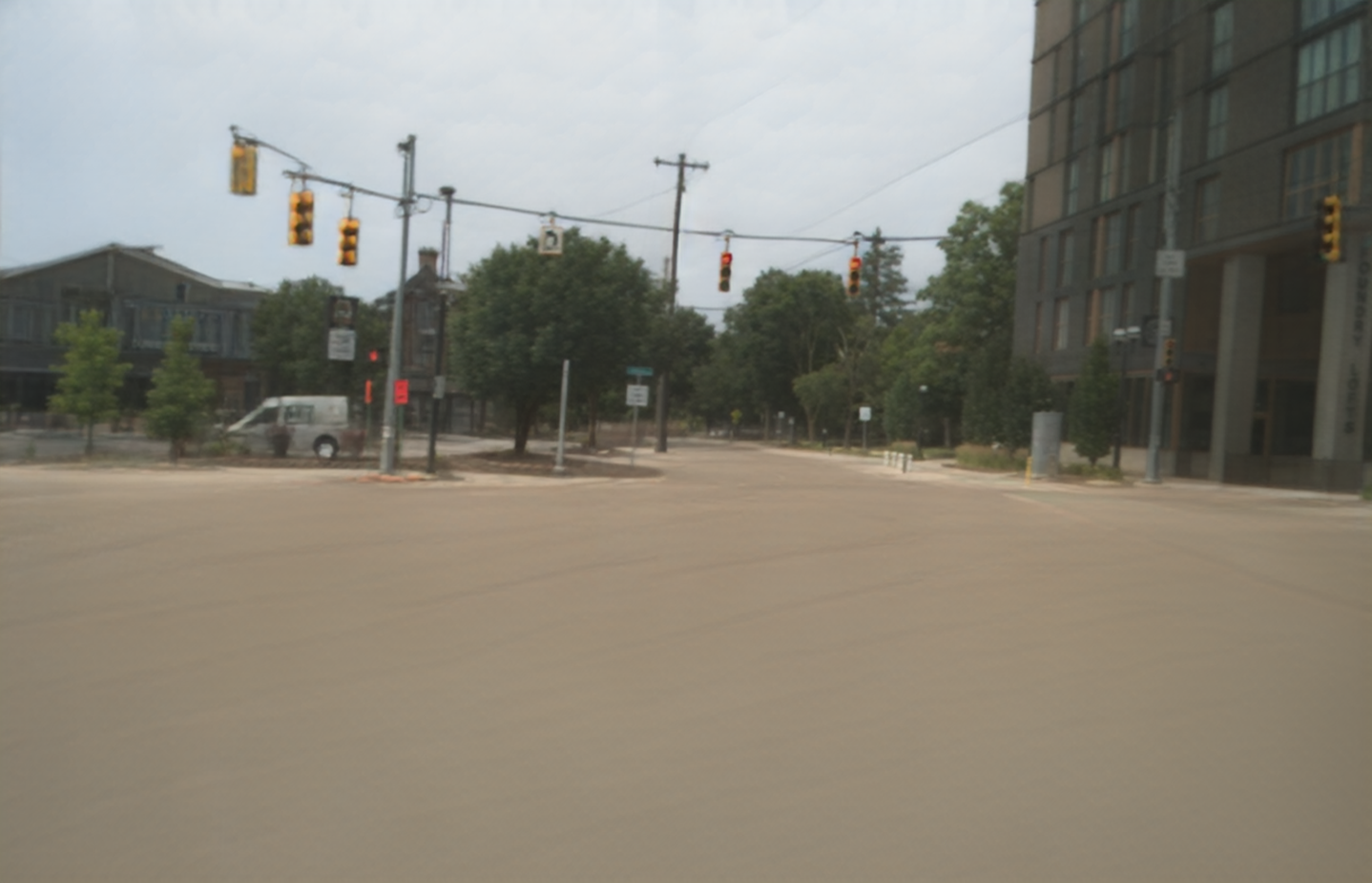}
\end{tabular}}
\caption{\textbf{Visual results of iterative 3DGS refinement.} Refining the 3DGS representation with $K=3$ refinement steps improves the final quality of the novel view render.}
\label{fig:iterative_refinement}
\end{figure}
\vspace{-0.5em}
\egroup

\section{Results}
\paragraph{Implementation and evaluation details.}
We train our 3DGS models at full resolution for 30k iterations, by minimizing an L1 and SSIM objective.
As for the enhancer, we finetune it from the public checkpoint~\cite{difix}  for 40k iterations with a learning rate of $2e-5$ without any schedule.
We report our numbers and ablation results either on the challenge data itself (\ie computed by the evaluation server) or on a validation set we built for development, composed of 6 Paralane (for Level 1 and 2) and 3 EUVS sequences (for Level 3). 
We will refer to the latter simply as the \emph{devset}.
Unless otherwise specified, we use the enhancer model in a single step (\ie without iterative 3DGS reconstruction).
\paragraph{3DGS improvements.}
As stated in \cref{sec:3dgs}, we found the initialization scheme to impact the 3DGS performance the most.
We report our study in \cref{tab:initialization}.
Whereas the challenge sequences come with their precomputed point clouds, simply applying the basic COLMAP pipeline yields $\approx 6\times$ more points, that boost the overall score from 0.319 to 0.343. 
Interestingly, the impact of the point cloud size on the 3DGS extrapolated views seems to saturate, and going for dense reconstruction mainly improves training view performance, with limited impact on the test views.
\paragraph{Enhancer results.}
\cref{tab:enhancer} shows the quantitative impact of the enhancer model on the devset sequences.
As compared to the pretrained model~\cite{difix}, finetuning the enhancer on 3DGS rendering artifacts in street scenes improves the overall score from 0.423 to 0.455, testifying for the effectiveness of our data curation strategy.
We also find that increasing the training resolution further improves the score up to 0.462.
Lastly, removing dynamic objects from the loss computation does not improve the overall score, yet it seems to benefit PSNR and SSIM.

In our experiments, we found the iterative 3DGS reconstruction strategy to improve the visual quality of the rendered frames (\cref{fig:iterative_refinement}), but such improvements are somehow not reflected in quantitative metrics.
We therefore excluded iterative 3DGS reconstruction from the final submission.

\paragraph{Qualitative results.}
We show qualitative results in \cref{fig:qualitative}, including the target view, renderings from the 3DGS backend, and enhanced images. 
The last row shows the main failure mode: extrapolating to entirely different trajectories (e.g. a perpendicular one in an intersection) remains a significant challenge, as the enhancer model will struggle to improve failed renders from the 3DGS.

\balance
{
    \small
    \bibliographystyle{ieeenat_fullname}
    \bibliography{bibliography}
}

\end{document}